\documentclass[journal]{IEEEtran}
\usepackage{graphicx}
\usepackage{graphics}
\usepackage{epsfig}
\usepackage{epstopdf}
\usepackage{stfloats}

\usepackage{xcolor,soul,framed} 

\colorlet{shadecolor}{yellow}
\usepackage[cmex10]{amsmath}
\usepackage{array}
\usepackage{mdwmath}
\usepackage{mdwtab}
\usepackage{eqparbox}
\usepackage{url}
\usepackage[colorlinks,
            linkcolor=red, 
             anchorcolor=blue, 
            citecolor=green, 
            ]{hyperref}
\usepackage{float}
\usepackage[ruled,linesnumbered,vlined]{algorithm2e}
\usepackage{algorithmic}
\usepackage{mathrsfs}
\usepackage{amssymb}
\usepackage{amsthm}
\usepackage{cite}
\usepackage{url}
\usepackage{xcolor}
\usepackage{subfig}
\usepackage{fancyhdr}
\usepackage{mdwtab}
\usepackage{caption}
\usepackage{stfloats}
\usepackage{array}
\usepackage{bbding}
\usepackage{subfloat}
\usepackage{verbatim}
\usepackage{threeparttable}

\usepackage[normalem]{ulem}
\usepackage{makecell}

\begin{document}
\bstctlcite{IEEEexample:BSTcontrol}
    \title{FBD-SV-2024: Flying Bird Object Detection Dataset in Surveillance Video}
  \author{Zi-Wei Sun,
          Ze-Xi Hua,
      Heng-Chao~Li,~\IEEEmembership{Senior Member,~IEEE},
      Zhi-Peng Qi,
      Xiang Li,
      Yan Li, and Jin-Chi Zhang

  \thanks{Manuscript completed 2024.  \emph{(Coressponding author: Ze-Xi Hua and Heng-Chao Li.)}}
  }

\markboth{ARXIV
}{Ziwei \MakeLowercase{\textit{et al.}}: FBD-SV-2024: Flying Bird Object Detection Dataset in Surveillance Video}

\maketitle

\begin{abstract}
A Flying Bird Dataset for Surveillance Videos (FBD-SV-2024) is introduced and tailored for the development and performance evaluation of flying bird detection algorithms in surveillance videos. This dataset comprises 483 video clips, amounting to 28,694 frames in total. Among them, 23,833 frames contain 28,366 instances of flying birds. The proposed dataset of flying birds in surveillance videos is collected from realistic surveillance scenarios, where the birds exhibit characteristics such as inconspicuous features in single frames (in some instances), generally small sizes, and shape variability during flight. These attributes pose challenges that need to be addressed when developing flying bird detection methods for surveillance videos. Finally, advanced (video) object detection algorithms were selected for experimentation on the proposed dataset, and the results demonstrated that this dataset remains challenging for the algorithms above. The FBD-SV-2024 is now publicly available: Please visit \href{https://github.com/Ziwei89/FBD-SV-2024_github}{https://github.com/Ziwei89/FBD-SV-2024\_github} for the dataset download link and related processing scripts.
\end{abstract}


%
\IEEEpeerreviewmaketitle


\section{Introduction}

\IEEEPARstart{B}{IRDS}, a common animal in our daily lives, can sometimes pose hazards to human production and living activities. Examples include bird strikes at airports \cite{2018_Tianhuang_skeleton_flying_bird}, birds pecking at crops \cite{2021_A_review_of_the_scientific_evidence}, and bird droppings causing short circuits and trips in substations \cite{2023_RCVNet}. These bird-related incidents seriously threaten the rapid socio-economic development. In response to these hazards, bird repelling is necessary. Traditional bird repelling methods involve continuously emitting physical signals (such as sound and light) to stimulate birds, thereby achieving the purpose of driving them away. These methods may be effective initially, but birds will gradually adapt to these physical stimuli, leading to a decline in the effectiveness of bird repelling \cite{2023_RCVNet, 2022_Two-Phase_Sensor}, which has prompted the development of the "detection-repelling" model for flying bird objects, in which the detection of flying bird objects is one of the primary tasks.

Initially, the detection of flying bird objects primarily relied on radar detection methods \cite{2016_Hoffmann_multistatic_radar, 2017_Jahangirstaring_radar}. Radar detection for flying birds offers advantages such as high detection accuracy, long detection range, and minimal interference from weather conditions. Nevertheless, radar detection also has numerous drawbacks \cite{2018_Tianhuang_skeleton_flying_bird, 2023_Li_Long-Distance_Avian_Identification}, including unfriendliness to human observation, making it inconvenient for humans to review, and the high cost of radar equipment, which hinders its widespread adoption in areas such as bird prevention in power grids and agriculture. In comparison to radar, the use of surveillance cameras for flying bird object detection based on computer vision offers the advantages of low cost, ease of deployment, and convenient maintenance. With the advancement of deep learning, object detection algorithms based on computer vision have achieved increasingly higher recognition accuracy \cite{yolov5_2021, 2021_Zheng_YOLOX, 2022_Chuyi_yolov6, 2023_Wang_yolov7, 2023_Guang_yolov8, 2024_wang_yolov9, 2024_wang_yolov10}, surpassing human vision in some domains. Among these, datasets (paired inputs and desired outputs) are crucial for the development of deep learning algorithms. Consequently, the Flying Bird Dataset for Surveillance Videos holds significant importance for the development of flying bird object detection algorithms in surveillance videos.

Currently, several datasets exist for bird objects. Yet, due to differing tasks or scenarios, the flying bird objects within them do not possess the characteristics commonly seen in surveillance videos (where, in most cases, single-frame image features are inconspicuous, object sizes are small, and shapes vary significantly during flight), rendering them unsuitable for the development of flying bird detection algorithms specifically for surveillance videos. Examples include the CUB-200-2011 dataset \cite{2011_Caltech-UCSD}, the Birdsnap dataset \cite{2014_Birdsnap}, and the NABirds dataset \cite{2015_NABirds}, where bird objects are typically large and clearly visible, making them primarily useful for bird species classification. The Drone2021 bird dataset proposed by Sanae Fujii et al. \cite{2021_Fujii_Distant_Bird_Detection_Drone} in 2021 and the MVA2023 dataset for flying birds introduced by Yuki Kondo et al. \cite{2023_MVA} in 2023, captured from a drone's perspective, can be applied in applications such as drone collision avoidance and bird dispersal but do not fully align with the needs of surveillance video detection. The wind farm flying bird dataset established by R. Yoshihashi et al. \cite{2017_Yoshihashi_Bird_detection_and_species} in 2015 and the airport flying bird dataset created by Hongyu Sun and his team \cite{2022_AirBirds} in 2022, though captured under surveillance camera angles, suffer from limited scene diversity and simplistic backgrounds, where flying bird objects are conspicuously featured. In summary, there remains a relative scarcity of datasets that can comprehensively reflect the characteristics of flying bird objects in surveillance videos, which in turn constrains the advancement of detection methods in this field.

Addressing the issues above, this paper presents a surveillance video flying bird object dataset (FBD-SV-2024) aimed at facilitating the development and performance evaluation of flying bird object detection algorithms for surveillance videos. This dataset comprises 483 video clips, totaling 28,694 frames of images. Among them, 23,833 frames contain 28,366 flying bird objects. A random selection of 83 video clips from this dataset serves as the test set, while the remaining 400 video clips constitute the training set. The video images in this dataset were captured in real-world surveillance scenarios, making them suitable for the development and performance evaluation of practical flying bird object detection algorithms for surveillance videos.

The structure of the remainder of this paper is as follows: Section \ref{The Proposed Dataset} introduces the proposed FBD-SV-2024 in detail. Section \ref{Experiment} conducts experiments using mainstream object detection algorithms on the proposed FBD-SV-2024. Section \ref{Conclusion} summarizes the work presented in this paper.

\section{The Proposed FBD-SV-2024}\label{The Proposed Dataset}

This section introduces the collected and curated FBD-SV-2024 dataset in two parts. Firstly, the creation process of the dataset, including its acquisition and annotation, is presented. Secondly, the characteristics of the flying bird objects within the dataset are analyzed.

\subsection{Collection and Annotation of the Dataset}\label{section_a}

The creation process of the FBD-SV-2024 dataset mainly consists of three steps:
\begin{itemize}
\item{Collection of video clips}
\item{Frame extraction from the videos}
\item{Annotation of flying bird objects}
\end{itemize}
The following provides a detailed account of these three steps.

\subsubsection{Collection of Video Clips}

Surveillance cameras (with a resolution of 1280×720 and a frame rate of 25fps) were deployed outdoors to capture surveillance videos over a period from December 2023 to May 2024. After acquiring the surveillance videos, through manual screening, video segments that featured flying bird objects were cut out (note that some frames within the video segments may still not contain flying bird objects). In the end, 483 video clips containing flying bird objects were collected, totaling 28,694 image frames. The collected video clips were numbered and named sequentially from 1 to 483 (bird\_1.mp4, bird\_2.mp4, ..., bird\_483.mp4).

\subsubsection{Frame Extraction From the Videos}

For the subsequent annotation of flying birds and the convenience of model training, the collected video clips are divided into video frames. Two naming conventions were employed for the video frames. The first involved appending the video name and sequentially numbering the frames starting from 0 (e.g., bird\_1\_000000.jpg, bird\_1\_000001.jpg, ...). This naming convention is convenient for image-based object detection methods. The second convention directly numbers the frames starting from 0. It organizes them within folders named after the corresponding video (e.g., bird\_1/000000.jpg, bird\_1/000001.jpg, ..., with all frames from the same video placed in a folder named after that video). This naming convention is advantageous for video-based object detection methods.

\subsubsection{Annotation of Flying Bird Objects}

The open-source tool ``labelImg"\footnote{Tzutalin. LabelImg. Git code (2015). \href{https://github.com/HumanSignal/labelImg}{https://github.com/HumanSignal/labelImg}} was utilized to annotate the categories (bird) and bounding boxes of flying bird objects in the images. The annotation process was conducted in three rounds, with each round handled by different individuals in a cross-checking manner (i.e., one person processed a batch of data in a particular round, and another handled the same batch in the subsequent round). The first round involved general annotation, the second round focused on checking for missed or incorrect annotations, and the third round refined the bounding boxes. Upon completion of annotation, two additional pieces of information were added to the labels: the object difficulty level and the object's ID within the current video. The difficulty level represents the recognition complexity, categorizing all flying bird objects into four levels ranging from 0 to 3, with 0 indicating easy, 1 indicating moderate, 2 indicating difficult, and 3 indicating hard (determined subjectively by humans, thus serving as a reference only). Additionally, video frames without flying bird objects were labeled, but these labels only contained basic image information (such as image name, image size, etc.) and did not include information related to flying bird objects. TABLE \ref{tab:Label_Information} summarizes the primary information of the label files used for object detection and video object detection, respectively.

\begin{table}[!ht]
\caption{Main Information in the Label File.\label{tab:Label_Information}}
\centering
\begin{tabular}{c|c c}
\hline
  & \makecell[c]{For Object\\Detection} & \makecell[c]{For Video\\ Object Detection}\\
\hline \hline
Image size  & width, height, depth & width, height\\
Categories  & bird & n01503061\\
Difficulty level  & difficult & difficult\\
Object ID  & - & track\_id\\
Bounding box  & xmin, ymin, xmax, ymax & xmin, ymin, xmax, ymax\\
\hline
\end{tabular}
\end{table}

After the annotation was completed, 83 video clips were randomly selected as the test set, and the remaining 400 video clips were used as the training set. TABLE \ref{tab:Train_and_val_Info} provides specific information for the training and test sets.

\begin{table}[!ht]
\caption{Specific Information of Training Set and Test Set.\label{tab:Train_and_val_Info}}
\centering
\begin{tabular}{c|c c c}
\hline
  & \makecell[c]{Training\\Set} &\makecell[c]{Test\\Set} & Total \\
\hline \hline
Number of videos  & 400 & 83 & 483\\
Number of images  & 23979 & 4715 & 28694\\
Number of images containing birds  & 19941 & 3892 & 23833\\
Number of easy object instances  & 2665 & 589 & 3254\\
Number of moderate object instances  & 12758 & 1949 & 14707\\
Number of difficult object instances  & 6983 & 1038 & 8021\\
Number of hard object instances  & 1633 & 751 & 2384\\
Total number of object instances  & 24039 & 4327 & 28366\\
\hline
\end{tabular}
\end{table}

\subsection{Characteristics of Flying Bird Objects in FBD-SV-2024 Dataset}\label{section_b}

Through observation and analysis, it is found that the bird objects in the collected dataset have some characteristics, such as inconspicuous features in single-frame images under certain circumstances, generally small sizes, and variable shapes during flight. Next, this article will analyze and present these characteristics one by one.

\subsubsection{In Some cases, the Features of Flying Bird Objects in Single-frame Images are not Obvious}

\begin{figure*}[!htp]
    \centering
    \subfloat[Examples of flying bird objects annotated with difficulty level 2]{
        \begin{minipage}[t]{0.9\linewidth}
        \centering
        \includegraphics[width=1\linewidth]{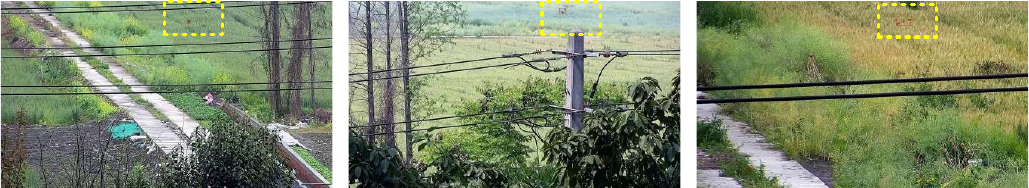}
        \label{Difficulty2}
        \scriptsize{$49^\text{{th}}$ frame frame of video bird\_122~~~~~~~~~~~~~~~~~~~$22^\text{{nd}}$ frame frame of video bird\_234~~~~~~~~~~~~~~~~~~~$50^\text{{th}}$ frame frame of video bird\_284}
        \end{minipage}
        }
    
    \vspace{-1mm}
    \subfloat[Examples of flying bird objects annotated with difficulty level 3]{
        \begin{minipage}[t]{0.9\linewidth}
        \centering
        \includegraphics[width=1\linewidth]{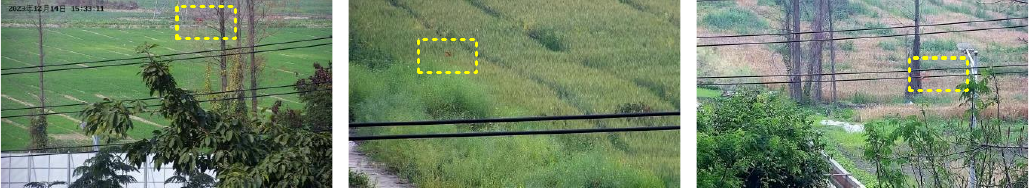}
        \label{Difficulty3}
        \scriptsize{$39^\text{{th}}$ frame frame of video bird\_14~~~~~~~~~~~~~~~~~~~$10^\text{{th}}$ frame frame of video bird\_212~~~~~~~~~~~~~~~~~~~$29^\text{{th}}$ frame frame of video bird\_327}
        \end{minipage}
        }
    \caption{Examples of flying bird objects with inconspicuous features in single-frame images (red boxes indicate the bounding boxes of the flying bird objects).}
    \label{Difficulty_diagram}
\end{figure*}

Due to the complexity of the background environment in surveillance videos, some flying bird objects (in single-frame images) blend seamlessly into the environment. In such cases, the human eye struggles to detect these flying bird objects solely based on a single frame. Statistics reveal that in the dataset, cases where the features of single-frame flying bird objects are not apparent (with a manual judgment difficulty level of 2 or 3) account for 36.7\% of all flying bird objects. Fig. \ref{Difficulty_diagram} showcases some examples of such flying bird objects.

\begin{figure*}[!htp]
    \centering
    \subfloat{
        \rotatebox{90}{\scriptsize{~~~~~bird\_122}}
        \begin{minipage}[t]{0.9\linewidth}
        \centering
        \includegraphics[width=1\linewidth]{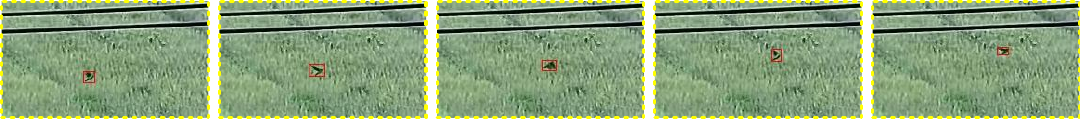}
        \label{bird_122}
        \scriptsize{$49^\text{{th}}$ frame~~~~~~~~~~~~~~~~~~~~~~~~~~$50^\text{{th}}$ frame~~~~~~~~~~~~~~~~~~~~~~~~~~$51^\text{{st}}$ frame~~~~~~~~~~~~~~~~~~~~~~~~~~$52^\text{{nd}}$ frame~~~~~~~~~~~~~~~~~~~~~~~~~~$53^\text{{rd}}$ frame}
        \end{minipage}
        }
    
    \vspace{-1mm}
    \subfloat{
        \rotatebox{90}{\scriptsize{~~~~~bird\_234}}
        \begin{minipage}[t]{0.9\linewidth}
        \centering
        \includegraphics[width=1\linewidth]{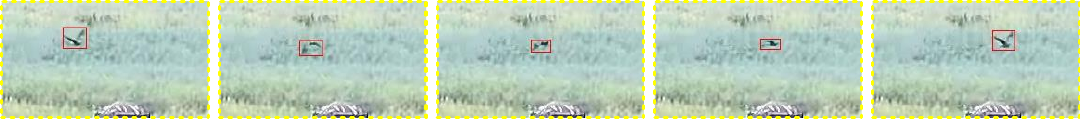}
        \scriptsize{$22^\text{{nd}}$ frame~~~~~~~~~~~~~~~~~~~~~~~~~~$23^\text{{rd}}$ frame~~~~~~~~~~~~~~~~~~~~~~~~~~$24^\text{{th}}$ frame~~~~~~~~~~~~~~~~~~~~~~~~~~$25^\text{{th}}$ frame~~~~~~~~~~~~~~~~~~~~~~~~~~$26^\text{{th}}$ frame}
        \end{minipage}
        }

    \vspace{-1mm}
    \subfloat{
        \rotatebox{90}{\scriptsize{~~~~~bird\_284}}
        \begin{minipage}[t]{0.9\linewidth}
        \centering
        \includegraphics[width=1\linewidth]{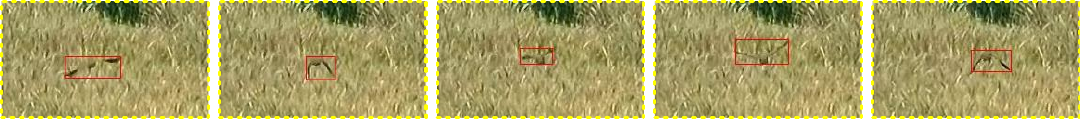}
        \scriptsize{$50^\text{{th}}$ frame~~~~~~~~~~~~~~~~~~~~~~~~~~$51^\text{{st}}$ frame~~~~~~~~~~~~~~~~~~~~~~~~~~$52^\text{{nd}}$ frame~~~~~~~~~~~~~~~~~~~~~~~~~~$53^\text{{rd}}$ frame~~~~~~~~~~~~~~~~~~~~~~~~~~$54^\text{{th}}$ frame}
        \end{minipage}
        }

    \vspace{-1mm}
    \subfloat{
        \rotatebox{90}{\scriptsize{~~~~~bird\_14}}
        \begin{minipage}[t]{0.9\linewidth}
        \centering
        \includegraphics[width=1\linewidth]{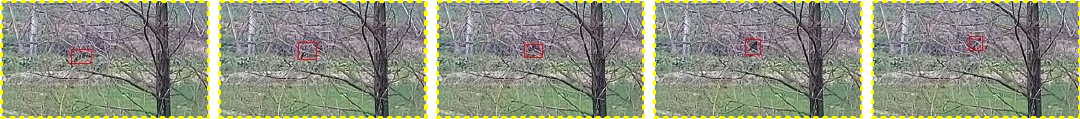}
        \scriptsize{$39^\text{{th}}$ frame~~~~~~~~~~~~~~~~~~~~~~~~~~$40^\text{{th}}$ frame~~~~~~~~~~~~~~~~~~~~~~~~~~$41^\text{{st}}$ frame~~~~~~~~~~~~~~~~~~~~~~~~~~$42^\text{{nd}}$ frame~~~~~~~~~~~~~~~~~~~~~~~~~~$43^\text{{rd}}$ frame}
        \end{minipage}
        }
    \vspace{-1mm}
    \subfloat{
        \rotatebox{90}{\scriptsize{~~~~~bird\_212}}
        \begin{minipage}[t]{0.9\linewidth}
        \centering
        \includegraphics[width=1\linewidth]{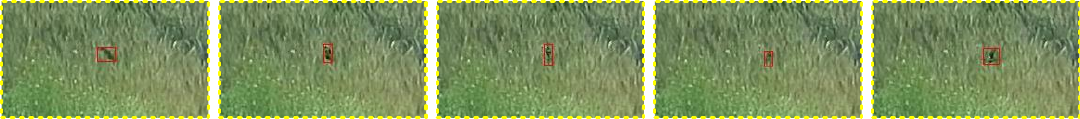}
        \scriptsize{$10^\text{{th}}$ frame~~~~~~~~~~~~~~~~~~~~~~~~~~$11^\text{{th}}$ frame~~~~~~~~~~~~~~~~~~~~~~~~~~$12^\text{{th}}$ frame~~~~~~~~~~~~~~~~~~~~~~~~~~$13^\text{{th}}$ frame~~~~~~~~~~~~~~~~~~~~~~~~~~$14^\text{{th}}$ frame}
        \end{minipage}
        }
    
    \vspace{-1mm}
    \subfloat{
        \rotatebox{90}{\scriptsize{~~~~~bird\_327}}
        \begin{minipage}[t]{0.9\linewidth}
        \centering
        \includegraphics[width=1\linewidth]{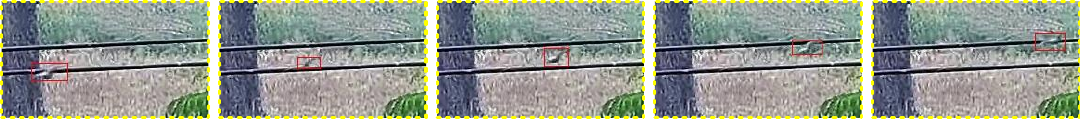}
        \scriptsize{$29^\text{{th}}$ frame~~~~~~~~~~~~~~~~~~~~~~~~~~$30^\text{{th}}$ frame~~~~~~~~~~~~~~~~~~~~~~~~~~$31^\text{{st}}$ frame~~~~~~~~~~~~~~~~~~~~~~~~~~$32^\text{{nd}}$ frame~~~~~~~~~~~~~~~~~~~~~~~~~~$33^\text{{rd}}$ frame}
        \end{minipage}
        }
    \caption{Screenshots of the corresponding flying bird objects in Fig. \ref{Difficulty_diagram} across 5 consecutive frames (highlighted by the yellow dashed box in Fig. \ref{Difficulty_diagram}).}
    \label{5_consecutive_frames}
\end{figure*}

Although the features of flying bird objects in single-frame images are not evident in these cases, observing consecutive frames can still reveal the presence of these objects, as shown in Fig. \ref{5_consecutive_frames}. Therefore, when training a flying bird object detection model using the dataset provided in this paper, it is recommended to consider the information of flying bird objects across consecutive frames.

\subsubsection{The Size of Flying Bird Objects is Generally Small}

\begin{figure}[htb]
\centering
\subfloat[]{\includegraphics[width=3.2in]{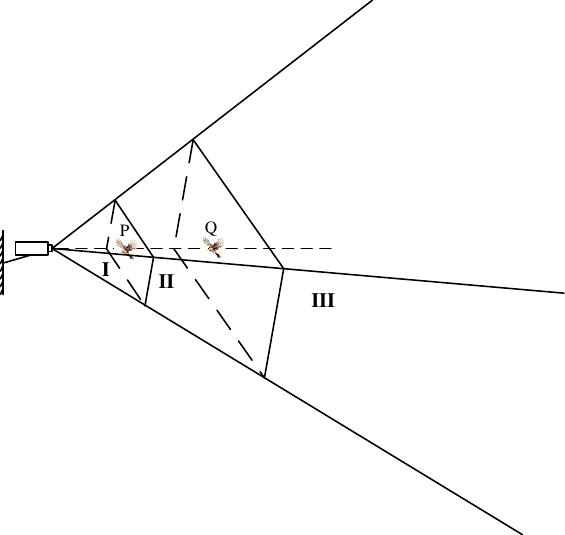}
\label{size_diagram_a}}
\\
\subfloat[]{\includegraphics[width=1.6in]{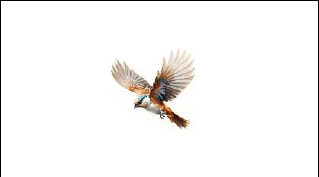}
\label{size_diagram_b}}
\subfloat[]{\includegraphics[width=1.6in]{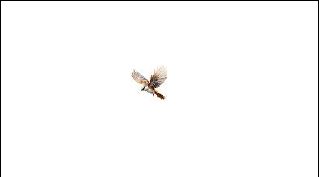}
\label{size_diagram_c}}
\caption{Schematic diagram of a surveillance camera capturing flying bird objects and the sizes of their imaging representations.}
\label{size_diagram}
\end{figure}

Fig. \ref{size_diagram}\subref{size_diagram_a} illustrates a schematic diagram of a surveillance camera capturing flying bird objects (in an ideal scenario without ground, perching sites, or obstructions). The figure shows that the camera's imaging space resembles a quadrangular pyramid. This paper divides this quadrangular pyramid space into three regions: I, II, and III. Flying bird objects located in Region I are closer to the camera, occupying more pixels in the video frame, and can be considered large objects [e.g., the flying bird object at point P, with its imaging shown in Fig. \ref{size_diagram}\subref{size_diagram_b}]. Flying bird objects in Regions II and III are farther from the camera, occupying fewer pixels in the video frame and appearing smaller compared to objects of the same size in Region I [e.g., the flying bird object at point Q, with its imaging shown in Fig. \ref{size_diagram}\subref{size_diagram_c}]. According to the properties of a quadrangular pyramid, the spatial volumes of Regions II and III are significantly larger than that of Region I. Therefore, if flying bird objects were uniformly distributed within the camera's imaging space (in reality, the distribution of flying bird objects in the camera's imaging space is influenced by various factors such as perching sites and obstructions; this assumption is made for a rough analysis of the size characteristics of flying bird objects in surveillance videos), the probability of objects being located in Regions II and III would be much higher than in Region I. Consequently, theoretically, flying bird objects in surveillance videos tend to be small in size.

\begin{figure*}[!ht]
\centering
\includegraphics[width=5in]{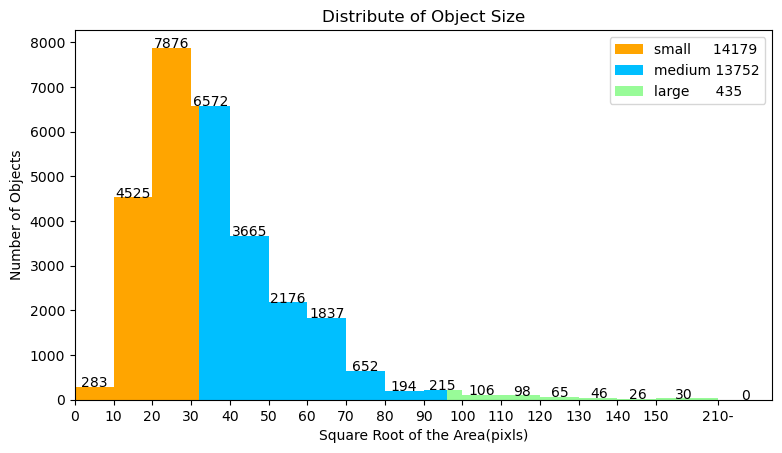}
\caption{Distribution of flying bird object sizes within the dataset.}
\label{size_curve_fig}
\end{figure*}

\begin{figure}[!ht]
\centering
\includegraphics[width=3.2in]{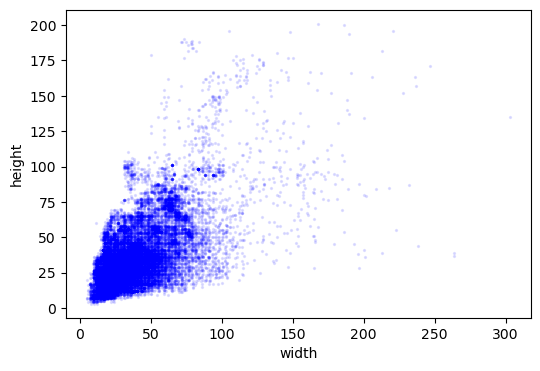}
\caption{Scatter plot of object sizes in the dataset.}
\label{scatter_plot}
\end{figure}

\begin{figure*}[!htp]
    \centering
    \subfloat{
        \begin{minipage}[t]{0.9\linewidth}
        \centering
        \includegraphics[width=1\linewidth]{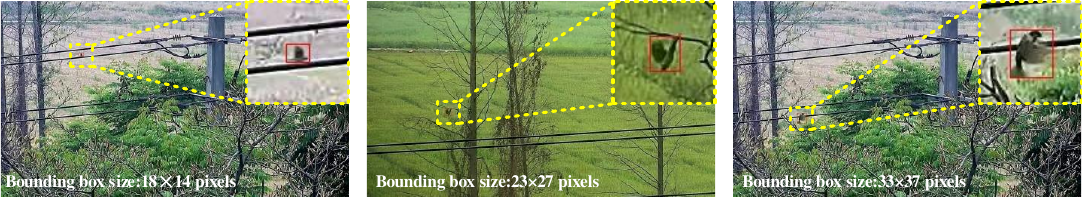}
        \scriptsize{$3^\text{{rd}}$ frame frame of video bird\_345~~~~~~~~~~~~~~~~~~~$33^\text{{rd}}$ frame frame of video bird\_148~~~~~~~~~~~~~~~~~~~$12^\text{{th}}$ frame frame of video bird\_364}
        \end{minipage}
        }
    
    \vspace{-1mm}
    \subfloat{
        \begin{minipage}[t]{0.9\linewidth}
        \centering
        \includegraphics[width=1\linewidth]{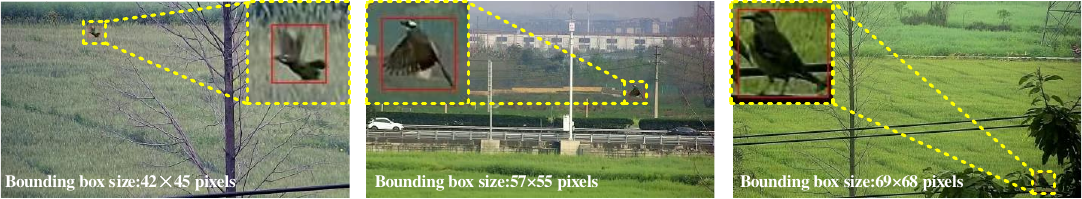}
        \scriptsize{$40^\text{{th}}$ frame frame of video bird\_173~~~~~~~~~~~~~~~~~~~$10^\text{{th}}$ frame frame of video bird\_116~~~~~~~~~~~~~~~~~~~$33^\text{{rd}}$ frame frame of video bird\_153}
        \end{minipage}
        }
    \caption{Examples of flying bird objects with sizes ranging from 10x10 to 70x70 pixels (the yellow dashed box before enlargement measures 80x80).}
    \label{small_size_diagram}
\end{figure*}

Furthermore, this paper also analyzes the size distribution of bird objects within the dataset (with Fig. \ref{size_curve_fig} presenting a bar chart of the distribution of bird object sizes and Fig .\ref{scatter_plot} showing a scatter plot of bird object sizes). From Fig. \ref{size_curve_fig}, it can be observed that the sizes of flying bird objects are primarily distributed within the range of 10×10 pixels to 70×70 pixels, accounting for approximately 94\% of all flying bird objects. Fig. \ref{small_size_diagram} showcases examples of flying bird objects from the dataset with sizes falling within the 10×10 to 70×70 pixel range. Statistical calculations reveal that flying bird objects with sizes smaller than 32×32 pixels constitute approximately 49.99\% of all objects, those with sizes between 32×32 and 96×96 pixels account for approximately 48.48\% and objects larger than 96×96 pixels make up only about 1.53\%. These statistics indicate that the dataset's flying bird objects are generally small. Therefore, special attention should be given to small-scale objects when developing methods for detecting flying bird objects in surveillance videos.

\subsubsection{The Appearance of Flying Birds Varies Greatly During Their Flight}

\begin{figure*}[!ht]
\centering
\includegraphics[width=6.4in]{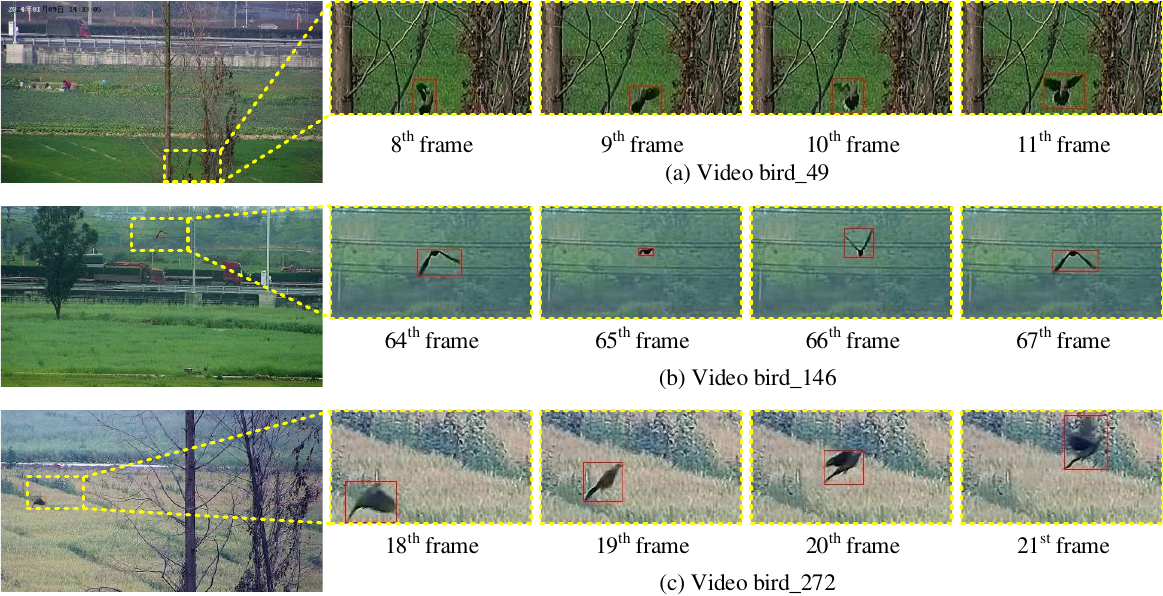}
\caption{Examples of flying bird objects in the process of flight captured by surveillance videos.}
\label{irregulate}
\end{figure*}

Most flying birds must flap their wings continuously to generate sufficient lift and maintain a balance between their gravitational and lift forces, which is precisely why flying birds are considered non-rigid objects, with their appearances constantly changing in a periodic (or non-periodic) manner. Fig. \ref{irregulate} demonstrates the various states of some flying bird objects during their flight in the dataset. As can be seen from the figure, the appearances of flying birds are in constant variation throughout their flight. Furthermore, the visual aspects of most flying bird objects are asymmetric and irregular in surveillance videos.

Due to the significant changes in the appearance of flying bird objects during flight, tasks related to the association of object bounding boxes between consecutive video frames (such as object tracking) should not overly rely on the Intersection over Union (IoU) between boxes when designing association matching algorithms. For the asymmetric and irregular appearance of flying bird objects, if an anchor-free deep learning approach is adopted for flying bird detection, it is recommended to employ a dynamic label assignment method when training the detection model.

\section{Experiment}\label{Experiment}

In this section, this paper adopts state-of-the-art object detection algorithms to conduct experiments on the proposed dataset, aiming to demonstrate that the dataset remains challenging even for these advanced algorithms. Simultaneously, it provides a comparative reference for developing flying bird object detection algorithms in surveillance videos utilizing the dataset proposed in this paper. Specifically, image-based object detection methods such as YOLOV5l \cite{yolov5_2021}, YOLOV6l \cite{2022_Chuyi_yolov6}, YOLOXl \cite{2021_Zheng_YOLOX}, YOLOV8l \cite{2023_Guang_yolov8}, YOLOV9e \cite{2024_wang_yolov9}, YOLOV10l \cite{2024_wang_yolov10}, SSD \cite{2016_Liu_SSD}, and video-based object detection methods like FGFA \cite{2017_Zhu_FGFA}, SELSA \cite{2019_Wu_SELSA}, Temporal RoI Align \cite{2021_Tao_Temporal_RoI_Align}, as well as our previous works FBOD-BMI \cite{2024_sun_Flying_Bird}, FBOD-SV \cite{2024_FBOD-SV} for flying bird object detection in surveillance videos, have been included in the experiments. Next, this paper will introduce and analyze the experimental platform (\ref{experimental_platforms}), implementation details (\ref{implementation_details}), evaluation methods (\ref{evaluation_metrics}), and experimental results (\ref{experimental_results}).

\subsection{Experimental Platforms}\label{experimental_platforms}

The experimental hardware platform is a desktop computer with an Intel Core i7-12700 CPU, 32 GB of RAM, and an NVIDIA GeForce RTX 3090 graphics card with 24 GB of video memory. The experimental software platform includes the Ubuntu 22.04 operating system, Python 3.10.6, Pytorch 1.11.0, and CUDA 11.3.

\subsection{Implementation Details}\label{implementation_details}

The (video) object detection algorithms such as YOLOV5l \cite{yolov5_2021}, YOLOV6l \cite{2022_Chuyi_yolov6}, YOLOXl \cite{2021_Zheng_YOLOX}, YOLOV8l \cite{2023_Guang_yolov8}, YOLOV9e \cite{2024_wang_yolov9}, YOLOV10l \cite{2024_wang_yolov10}, SSD \cite{2016_Liu_SSD}, FGFA \cite{2017_Zhu_FGFA}, SELSA \cite{2019_Wu_SELSA}, and Temporal RoI Align \cite{2021_Tao_Temporal_RoI_Align} utilized their respective open-source codes. Among them, YOLOV5l \cite{yolov5_2021}, YOLOV6l \cite{2022_Chuyi_yolov6}, YOLOV8l \cite{2023_Guang_yolov8}, YOLOV9e \cite{2024_wang_yolov9}, and YOLOV10l \cite{2024_wang_yolov10} employ the open-source framework\footnote{\href{https://github.com/ultralytics/ultralytics}{https://github.com/ultralytics/ultralytics}} provided by Ultralytics, while YOLOXl \cite{2021_Zheng_YOLOX} and SSD \cite{2016_Liu_SSD} utilize the mmdetection \cite{mmdetection} open-source framework. FGFA \cite{2017_Zhu_FGFA}, SELSA \cite{2019_Wu_SELSA}, and Temporal RoI Align \cite{2021_Tao_Temporal_RoI_Align} adopt the mmtracking \cite{mmtrack2020} open-source framework. All models were trained from scratch on the training set without using pre-trained models. All data augmentation methods provided by the corresponding open-source frameworks were applied during training. Evaluations were conducted on the test set.

\subsection{Evaluation Metrics}\label{evaluation_metrics}

This paper adopts the evaluation metric of average precision (AP) from Pascal VOC 2007 \cite{2010_Pascal_VOC}, a commonly used metric for evaluating object detection algorithms, to evaluate the detection results of the models.

\subsection{Experimental Results}\label{experimental_results}

The quantitative experimental results are shown in TABLE \ref{tab:experiment}. The detection accuracy AP50 of image-based object detection methods is generally higher than that of video-based object detection methods. The reason for this phenomenon has been analyzed in our previous work \cite{2024_sun_Flying_Bird, 2024_FBOD-SV}: Although video object detection methods utilize information from multiple frames, they extract features from single frames during the initial feature extraction stage, which can lead to the loss of weakly featured flying bird objects during the feature extraction process. During feature aggregation, erroneous features may be formed, reducing detection accuracy. Notably, our previous work FBOD-SV \cite{2024_FBOD-SV} achieved the highest detection accuracy AP50 of 71.9\%, which is 9.9\% higher than YOLOXl \cite{2021_Zheng_YOLOX} (the highest-performing image-based object detection method with an AP50 of 62.0\%) and 31.9\% higher than SELSA \cite{2019_Wu_SELSA} (the highest-performing video-based object detection method with an AP50 of 40.0\%). The main reason for the improved detection accuracy of FBOD-SV \cite{2024_FBOD-SV} compared to YOLOXl \cite{2021_Zheng_YOLOX} is that FBOD-SV \cite{2024_FBOD-SV} aggregates features from consecutive $n$ frames before extracting features for flying bird objects, thereby enhancing the features of flying bird objects in a single frame.

\begin{table*}[!ht]
\caption{Quantitative Experimental Results ($\text{AP}_\text{S}$, $\text{AP}_\text{M}$, $\text{AP}_\text{L}$ represent the detection performance of flying bird objects with sizes ranging from 0$\sim$32$\times$32 pixels, 32$\times$32$\sim$48$\times$48 pixels, and 48$\times$48$\sim$ pixels, respectively)\label{tab:experiment}}
\centering
\begin{threeparttable}
\begin{tabular}{c|c| c c c c c c}
\hline
Method  & Image Size  & $\text{AP}_{50}$ & $\text{AP}_{75}$  & AP & $\text{AP}_\text{S}$ & $\text{AP}_\text{M}$ & $\text{AP}_\text{L}$\\
\hline \hline
YOLOV5l \cite{yolov5_2021} & 640$\times$640 & 0.558 & 0.307 & 0.299 & 0.220 & 0.417 & 0.521\\
YOLOV6l \cite{2022_Chuyi_yolov6} & 640$\times$640 & 0.585 & 0.308 & 0.304 & 0.214 & 0.437 & 0.528\\
YOLOXl \cite{2021_Zheng_YOLOX} & 640$\times$640 & 0.620 & 0.337 & 0.338 & 0.280 & 0.423 & 0.561\\
YOLOV8l \cite{2023_Guang_yolov8} & 640$\times$640 & 0.584 & 0.327 & 0.318 & 0.233 & 0.437 & 0.597\\
YOLOV9e \cite{2024_wang_yolov9} & 640$\times$640 & 0.577 & 0.322 & 0.318 & 0.227 & 0.456 & 0.577\\
YOLOV10l \cite{2024_wang_yolov10} & 640$\times$640 & 0.550 & 0.289 & 0.294 & 0.199 & 0.424 & 0.598\\
SSD \cite{2016_Liu_SSD} & 640$\times$640 & 0.599 & 0.277 & 0.299 & 0.231 & 0.395 & 0.510\\
FGFA \cite{2017_Zhu_FGFA} & 1000$\times$600 & 0.198 & 0.072 & 0.092 & 0.020 & 0.187 & 0.420\\
SELSA \cite{2019_Wu_SELSA} & 1000$\times$600 & 0.400 & 0.162 & 0.193 & 0.050 & 0.414 & 0.614\\
Temporal RoI Align \cite{2021_Tao_Temporal_RoI_Align} & 1000$\times$600 & 0.371 & 0.180 & 0.186 & 0.038 & 0.411 & 0.582\\
FBOD-BMI \cite{2024_sun_Flying_Bird} & 672$\times$384 & 0.692 & 0.302 & 0.351 & 0.306 & 0.434 & 0.401\\
FBOD-SV \cite{2024_FBOD-SV} & 672$\times$384 & 0.719 & 0.341 & 0.371 & 0.313 & 0.479 & 0.425\\
\hline
\end{tabular}
\end{threeparttable}
\end{table*}

\begin{figure*}[!htp]
    \centering
    \subfloat[]{
        \begin{minipage}[t]{0.225\linewidth}
        \centering
        \includegraphics[width=1\linewidth]{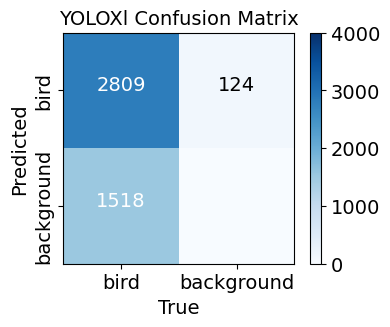}
        \label{yolox}
        \end{minipage}
        }
    \subfloat[]{
        \begin{minipage}[t]{0.225\linewidth}
        \centering
        \includegraphics[width=1\linewidth]{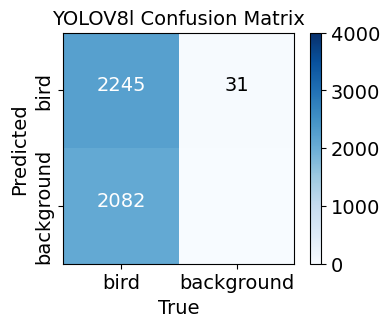}
        \label{yolov8}
        \end{minipage}
        }
    \subfloat[]{
        \begin{minipage}[t]{0.225\linewidth}
        \centering
        \includegraphics[width=1\linewidth]{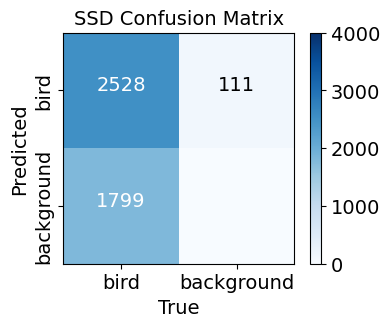}
        \label{ssd}
        \end{minipage}
        }
    \subfloat[]{
        \begin{minipage}[t]{0.225\linewidth}
        \centering
        \includegraphics[width=1\linewidth]{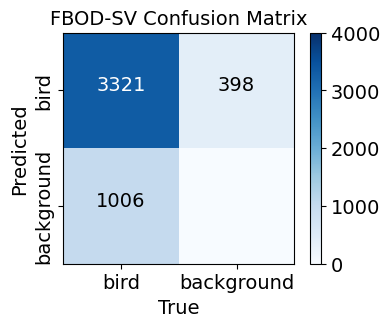}
        \label{fbod-sv}
        \end{minipage}
        }
    \caption{Confusion matrices of several advanced detection methods on the FBD-SV-2024 test set, where the True Positive IoU threshold is set to 0.2 and the True Positive score threshold is set to 0.3.}
    \label{confusion_matrices}
\end{figure*}

Furthermore, this paper selects several advanced detection methods (YOLOXl \cite{2021_Zheng_YOLOX}, YOLOV8l \cite{2023_Guang_yolov8}, SSD \cite{2016_Liu_SSD}, FBOD-SV \cite{2024_FBOD-SV}) and calculates their confusion matrices on the FBD-SV-2024 test set.  As illustrated in Fig. \ref{confusion_matrices}, the confusion matrices of these methods reveal their respective performance characteristics.  Specifically, YOLOXl \cite{2021_Zheng_YOLOX}, YOLOV8l \cite{2023_Guang_yolov8}, and SSD \cite{2016_Liu_SSD} exhibit relatively significant missed detections during testing, failing to identify some objects. In contrast, while FBOD-SV \cite{2024_FBOD-SV} has a slightly lower missed detection rate, it suffers from a higher false detection rate (i.e., misidentifying non-target objects as objects), generating 398 false detections.

It is clear from the above experimental results that current state-of-the-art video object detection methods still face challenges in detecting flying bird objects in surveillance videos, failing to achieve ideal detection performance. Specifically, the quantitative evaluation metrics are relatively low, indicating that there are prone to many false detections and missed detections during the prediction and inference process. Although the specially designed detection method FBOD-SV \cite{2024_FBOD-SV} for flying bird objects in surveillance videos performed best in the comparative experiments, its performance still needs to be improved as there are still a certain number of false detections and missed detections, which means that this method still has a particular gap from meeting the requirements of practical applications.



\section{Conclusion}\label{Conclusion}

This paper proposes a dataset (FBD-SV-2024) targeting flying bird objects in surveillance videos, aiming to facilitate the development and performance evaluation of flying bird detection algorithms for surveillance videos. The flying bird object in this dataset exhibits several notable characteristics, including inconspicuous features in single frames under certain conditions, generally small sizes, and varied shapes during flight. To validate the challenges posed by this dataset, we conducted experiments using current state-of-the-art object detection algorithms. The experimental results show that the dataset still presents a high challenge even for these advanced detection algorithms.


%





\ifCLASSOPTIONcaptionsoff
  \newpage
\fi





\bibliographystyle{IEEEtran}
\bibliography{IEEEabrv,Bibliography}

\begin{thebibliography}{10}
\providecommand{\url}[1]{#1}
\csname url@rmstyle\endcsname
\providecommand{\newblock}{\relax}
\providecommand{\bibinfo}[2]{#2}
\providecommand\BIBentrySTDinterwordspacing{\spaceskip=0pt\relax}
\providecommand\BIBentryALTinterwordstretchfactor{4}
\providecommand\BIBentryALTinterwordspacing{\spaceskip=\fontdimen2\font plus
\BIBentryALTinterwordstretchfactor\fontdimen3\font minus \fontdimen4\font\relax}
\providecommand\BIBforeignlanguage[2]{{%
\expandafter\ifx\csname l@#1\endcsname\relax
\typeout{** WARNING: IEEEtran.bst: No hyphenation pattern has been}%
\typeout{** loaded for the language `#1'. Using the pattern for}%
\typeout{** the default language instead.}%
\else
\language=\csname l@#1\endcsname
\fi
#2}}

\bibitem{2018_Tianhuang_skeleton_flying_bird}
\BIBentryALTinterwordspacing
T.~WU, X.~LUO, and Q.~XU, ``A new skeleton based flying bird detection method for low-altitude air traffic management,'' \emph{Chinese Journal of Aeronautics}, vol.~31, no.~11, pp. 2149--2164, 2018. [Online]. Available: \url{https://www.sciencedirect.com/science/article/pii/S1000936118300360}
\BIBentrySTDinterwordspacing

\bibitem{2021_A_review_of_the_scientific_evidence}
J.~K. Enos, M.~P. Ward, and M.~E. Hauber, ``A review of the scientific evidence on the impact of biologically salient frightening devices to protect crops from avian pests,'' \emph{Crop Protection}, vol. 148, p. 105734, 2021.

\bibitem{2023_RCVNet}
W.~Gao, Y.~Wu, C.~Hong, R.-J. Wai, and C.-T. Fan, ``Rcvnet: A bird damage identification network for power towers based on fusion of rf images and visual images,'' \emph{Advanced Engineering Informatics}, vol.~57, p. 102104, 2023.

\bibitem{2022_Two-Phase_Sensor}
T.~Cinkler, K.~Nagy, C.~Simon, R.~Vida, and H.~Rajab, ``Two-phase sensor decision: Machine-learning for bird sound recognition and vineyard protection,'' \emph{IEEE Sensors Journal}, vol.~22, no.~12, pp. 11\,393--11\,404, 2022.

\bibitem{2016_Hoffmann_multistatic_radar}
F.~Hoffmann, M.~Ritchie, F.~Fioranelli, A.~Charlish, and H.~Griffiths, ``Micro-doppler based detection and tracking of uavs with multistatic radar,'' in \emph{2016 IEEE Radar Conference (RadarConf)}, 2016, pp. 1--6.

\bibitem{2017_Jahangirstaring_radar}
M.~Jahangir, C.~J. Baker, and G.~A. Oswald, ``Doppler characteristics of micro-drones with l-band multibeam staring radar,'' in \emph{2017 IEEE Radar Conference (RadarConf)}, 2017, pp. 1052--1057.

\bibitem{2023_Li_Long-Distance_Avian_Identification}
J.~Li, K.~Shimasaki, and I.~Ishii, ``Long-distance avian identification approach based on high-frame-rate video,'' in \emph{2023 IEEE 19th International Conference on Automation Science and Engineering (CASE)}, 2023, pp. 1--7.

\bibitem{yolov5_2021}
Y.~Contributors, ``You only look once version 5,'' \url{https://github.com/ultralytics/yolov5}, 2021.

\bibitem{2021_Zheng_YOLOX}
\BIBentryALTinterwordspacing
Z.~Ge, S.~Liu, F.~Wang, Z.~Li, and J.~Sun, ``{YOLOX:} exceeding {YOLO} series in 2021,'' \emph{CoRR}, vol. abs/2107.08430, 2021. [Online]. Available: \url{https://arxiv.org/abs/2107.08430}
\BIBentrySTDinterwordspacing

\bibitem{2022_Chuyi_yolov6}
C.~Li, L.~Li, H.~Jiang, K.~Weng, Y.~Geng, L.~Li, Z.~Ke, Q.~Li, M.~Cheng, W.~Nie, Y.~Li, B.~Zhang, Y.~Liang, L.~Zhou, X.~Xu, X.~Chu, X.~Wei, and X.~Wei, ``Yolov6: A single-stage object detection framework for industrial applications,'' 2022.

\bibitem{2023_Wang_yolov7}
\BIBentryALTinterwordspacing
C.-Y. Wang, A.~Bochkovskiy, and H.-Y.~M. Liao, ``Yolov7: Trainable bag-of-freebies sets new state-of-the-art for real-time object detectors,'' \emph{2023 IEEE/CVF Conference on Computer Vision and Pattern Recognition (CVPR)}, Jun 2023. [Online]. Available: \url{http://dx.doi.org/10.1109/CVPR52729.2023.00721}
\BIBentrySTDinterwordspacing

\bibitem{2023_Guang_yolov8}
G.~J.~N. Ang, A.~K. Goil, H.~Chan, J.~J. Lew, X.~C. Lee, R.~B.~A. Mustaffa, T.~Jason, Z.~T. Woon, and B.~Shen, ``A novel application for real-time arrhythmia detection using yolov8,'' 2023.

\bibitem{2024_wang_yolov9}
C.-Y. Wang, I.-H. Yeh, and H.-Y.~M. Liao, ``Yolov9: Learning what you want to learn using programmable gradient information,'' 2024.

\bibitem{2024_wang_yolov10}
A.~Wang, H.~Chen, L.~Liu, K.~Chen, Z.~Lin, J.~Han, and G.~Ding, ``Yolov10: Real-time end-to-end object detection,'' \emph{arXiv preprint arXiv:2405.14458}, 2024.

\bibitem{2011_Caltech-UCSD}
C.~Wah, S.~Branson, P.~Welinder, P.~Perona, and S.~Belongie, ``Caltech-ucsd birds-200-2011,'' California Institute of Technology, Tech. Rep. CNS-TR-2011-001, 2011.

\bibitem{2014_Birdsnap}
T.~Berg, J.~Liu, S.~W. Lee, M.~L. Alexander, D.~W. Jacobs, and P.~N. Belhumeur, ``Birdsnap: Large-scale fine-grained visual categorization of birds,'' in \emph{2014 IEEE Conference on Computer Vision and Pattern Recognition}, 2014, pp. 2019--2026.

\bibitem{2015_NABirds}
G.~Van~Horn, S.~Branson, R.~Farrell, S.~Haber, J.~Barry, P.~Ipeirotis, P.~Perona, and S.~Belongie, ``Building a bird recognition app and large scale dataset with citizen scientists: The fine print in fine-grained dataset collection,'' in \emph{2015 IEEE Conference on Computer Vision and Pattern Recognition (CVPR)}, 2015, pp. 595--604.

\bibitem{2021_Fujii_Distant_Bird_Detection_Drone}
S.~Fujii, K.~Akita, and N.~Ukita, ``Distant bird detection for safe drone flight and its dataset,'' in \emph{2021 17th International Conference on Machine Vision and Applications (MVA)}, 2021, pp. 1--5.

\bibitem{2023_MVA}
Y.~Kondo, N.~Ukita, T.~Yamaguchi, H.-Y. Hou, M.-Y. Shen, C.-C. Hsu, E.-M. Huang, Y.-C. Huang, Y.-C. Xia, C.-Y. Wang, C.-Y. Lee, D.~Huo, M.~A. Kastner, T.~Liu, Y.~Kawanishi, T.~Hirayama, T.~Komamizu, I.~Ide, Y.~Shinya, X.~Liu, G.~Liang, and S.~Yasui, ``Mva2023 small object detection challenge for spotting birds: Dataset, methods, and results,'' in \emph{2023 18th International Conference on Machine Vision and Applications (MVA)}, 2023, pp. 1--11.

\bibitem{2017_Yoshihashi_Bird_detection_and_species}
R.~Yoshihashi, R.~Kawakami, M.~Iida, and T.~Naemura, ``Bird detection and species classification with time-lapse images around a wind farm: Dataset construction and evaluation,'' \emph{Wind Energy}, vol.~20, no.~12, pp. 1983--1995, 2017.

\bibitem{2022_AirBirds}
H.~Sun, Y.~Wang, X.~Cai, P.~Wang, Z.~Huang, D.~Li, Y.~Shao, and S.~Wang, ``Airbirds: A large-scale challenging dataset for bird strike prevention in real-world airports,'' in \emph{Computer Vision -- ACCV 2022}, L.~Wang, J.~Gall, T.-J. Chin, I.~Sato, and R.~Chellappa, Eds.\hskip 1em plus 0.5em minus 0.4em\relax Cham: Springer Nature Switzerland, 2023, pp. 409--424.

\bibitem{2016_Liu_SSD}
W.~Liu, D.~Anguelov, D.~Erhan, C.~Szegedy, S.~Reed, C.~Y. Fu, and A.~C. Berg, ``Ssd: Single shot multibox detector,'' in \emph{2016 European Conference on Computer Vision (ECCV)}, 2016.

\bibitem{2017_Zhu_FGFA}
X.~Zhu, Y.~Wang, J.~Dai, L.~Yuan, and Y.~Wei, ``Flow-guided feature aggregation for video object detection,'' in \emph{2017 IEEE International Conference on Computer Vision (ICCV)}, 2017, pp. 408--417.

\bibitem{2019_Wu_SELSA}
H.~Wu, Y.~Chen, N.~Wang, and Z.-X. Zhang, ``Sequence level semantics aggregation for video object detection,'' in \emph{2019 IEEE/CVF International Conference on Computer Vision (ICCV)}, 2019, pp. 9216--9224.

\bibitem{2021_Tao_Temporal_RoI_Align}
T.~Gong, K.~Chen, X.~Wang, Q.~Chu, F.~Zhu, D.~Lin, N.~Yu, and H.~Feng, ``Temporal roi align for video object recognition,'' in \emph{The Thirty-Fifth AAAI Conference on Artificial Intelligence (AAAI-21)}, 2021, pp. 1442--1450.

\bibitem{2024_sun_Flying_Bird}
Z.-W. Sun, Z.-X. Hua, H.-C. Li, and H.-Y. Zhong, ``Flying bird object detection algorithm in surveillance video based on motion information,'' \emph{IEEE Transactions on Instrumentation and Measurement}, vol.~73, pp. 1--15, 2024.

\bibitem{2024_FBOD-SV}
Z.-W. Sun, Z.-X. Hua, H.-C. Li, and Y.~Li, ``A flying bird object detection method for surveillance video,'' \emph{IEEE Transactions on Instrumentation and Measurement}, vol.~73, pp. 1--14, 2024.

\bibitem{mmdetection}
K.~Chen, J.~Wang, J.~Pang, Y.~Cao, Y.~Xiong, X.~Li, S.~Sun, W.~Feng, Z.~Liu, J.~Xu, Z.~Zhang, D.~Cheng, C.~Zhu, T.~Cheng, Q.~Zhao, B.~Li, X.~Lu, R.~Zhu, Y.~Wu, J.~Dai, J.~Wang, J.~Shi, W.~Ouyang, C.~C. Loy, and D.~Lin, ``{MMDetection}: Open mmlab detection toolbox and benchmark,'' \emph{arXiv preprint arXiv:1906.07155}, 2019.

\bibitem{mmtrack2020}
M.~Contributors, ``{MMTracking: OpenMMLab} video perception toolbox and benchmark,'' \url{https://github.com/open-mmlab/mmtracking}, 2020.

\bibitem{2010_Pascal_VOC}
\BIBentryALTinterwordspacing
M.~Everingham, L.~Van~Gool, C.~K.~I. Williams, J.~Winn, and A.~Zisserman, ``The pascal visual object classes (voc) challenge,'' \emph{International Journal of Computer Vision}, vol.~88, no.~2, pp. 303--338, Jun 2010. [Online]. Available: \url{https://doi.org/10.1007/s11263-009-0275-4}
\BIBentrySTDinterwordspacing

\end{thebibliography}

\vfill


\end{document}